\documentclass[letterpaper]{article} 
\usepackage[preprint]{aaai2027}
\usepackage[hyphens]{url} 
\usepackage{graphicx} 
\usepackage{natbib} 
\usepackage{caption} 
\usepackage{algorithm}
\usepackage{algorithmic}
\usepackage{booktabs}
\usepackage{multirow}
\usepackage{amsmath}
\usepackage{amssymb}
\usepackage{array}

\usepackage{comment}

\title{VeriSkill: A Self-Evolution Framework for Program Verification Skills}

\author{
    Changguo Jia\textsuperscript{\rm 1}\equalcontrib,
    Tianqi Zhao\textsuperscript{\rm 2}\equalcontrib,
    Zhiyou Xiao\textsuperscript{\rm 1},
    Weiming Zhang\textsuperscript{\rm 3},
    Minghui Zhou\textsuperscript{\rm 1}\corresponding
}

\affiliations{
    \textsuperscript{\rm 1}Peking University, Beijing, China\\
    \textsuperscript{\rm 2}Zhongguancun Laboratory, Beijing, China\\
    \textsuperscript{\rm 3}Shanghai Jiao Tong University, Shanghai, China\\
    jiachangguo@stu.pku.edu.cn, zhaotq@zgclab.edu.cn, xiaozhiyou@stu.pku.edu.cn, \\ WeimingZhang\_2020@sjtu.edu.cn, zhmh@pku.edu.cn
}

\begin{document}

\maketitle

\begin{abstract}
Automating program verification with LLM agents requires generating specifications, annotations, auxiliary lemmas, and tool invocations, all of which depend on reusable skills. A natural remedy is skill self‑evolution: distilling skills from trajectories and refining them through feedback. However, existing evolution methods struggle with program verification tasks because they cannot reliably identify skill‑specific failures or extract actionable signals from opaque verifier feedback. 
In this paper, we propose \textsc{VeriSkill}, a self‑evolution framework built for program verification. It attributes verification failures to skill deficiencies, distills diagnostic signatures into reusable lessons, and iteratively refines candidate skills, admitting only revisions that improve verification performance while preserving program semantics. 
Experiments show that \textsc{VeriSkill} consistently outperforms all baselines across multiple verification tools, agent frameworks, and LLM backends.
\end{abstract}


\section{Introduction}

Program verification aims to establish that a program implementation satisfies its formal specification~\cite{hoare1969axiomatic}. 
In practice, users manually annotate programs with function contracts, intermediate assertions, and auxiliary lemmas, after which verification tools generate the corresponding verification
conditions and invoke Satisfiability Modulo Theories (SMT) solvers to discharge them. 
Constructing correct annotations requires considerable
expertise and manual effort, making program verification at
scale challenging. 
Recently, the emergence of LLM agents and agent skills has made fully automated program verification increasingly feasible.


Agent skills encapsulate reusable operational knowledge that enables LLM agents to perform specialized tasks. 
A skill typically contains task-specific workflows, tool-use instructions, and error-handling procedures. 
Empirical studies show that skill quality substantially affects agent behavior~\cite{li2026skillsbenchbenchmarkingagentskills,zhou2026skillgenbench}. 
Yet writing and maintaining a high-quality skill requires expert effort that is always in short supply.
\emph{Skill self-evolution} addresses this bottleneck by automatically revising skills based on accumulated task experience.


Most existing methods of skill self-evolution are domain-agnostic, extracting lessons from successful and failed execution trajectories and incorporating them into skills. 
Representative methods follow this trajectory-based paradigm in different ways. 
Trace2Skill distills trajectory-local experience into transferable instructions~\cite{ni2026trace2skill}. 
SkillOpt-Lite streamlines skill self-evolution into file-system-based trajectory exploration, consensus attribute mining, and independent validation gating~\cite{shen2026skillopt}, while 
EvoSkill analyzes execution failures to iteratively refine structured skills under validation-based frontier selection~\cite{alzubi2026evoskill}.

However, general methods of skill self-evolution are not well suited to program verification. 
A failed verification attempt does not necessarily indicate a skill deficiency, making it difficult to determine whether the skill should be revised. 
For example, even when the source program and its annotations are both correct, verification may fail because of limitations in the underlying SMT solver, which are unrelated to the skill. 
Moreover, outputs from verification tools report failed verification conditions, which are not directly reusable procedural knowledge. 
For example, an incorrect loop invariant often results from a flawed mathematical derivation, while the resulting failed verification conditions provide little guidance for correcting the derivation. 
As a result, program verification skills require a specialized self-evolution framework.

In this paper, we propose \textsc{VeriSkill}, a self-evolution framework 
for program verification skills that turns verification failures into actionable improvements. 
Unlike generic evolution approaches, \textsc{VeriSkill} precisely isolates skill-related failures, extracts generalizable lessons while discarding instance-specific noise, and admits only validated lessons that measurably boost verification success without breaking program semantics—making skill self-evolution both targeted and trustworthy.
The framework 
operates
in three steps. 
First, to determine whether a failed verification attempt indicates a skill deficiency, \textsc{VeriSkill} performs responsibility attribution for each verification failure by examining the failed task, verifier feedback, current skill, generated artifact, human-verified artifact, and evolution memory. It distinguishes failures and retains only the skill-responsible failures as evolution signals. 
Second, to turn failed verification conditions into reusable procedural knowledge, \textsc{VeriSkill} clusters failures by diagnostic signatures and abstracts a reusable lesson from each failure pattern. Each signature captures where the verification attempt fails, what kind of skill deficiency is exposed, and what procedural guidance is missing. Human-verified artifacts are used as evidence for the missing guidance, but instance-specific verification details are removed so that the resulting lesson specifies general applicability conditions and non-applicable cases. 
Finally, \textsc{VeriSkill} iteratively refines candidate skills through executable validation. Each lesson is integrated into a candidate skill and revised on same-pattern attribution and transfer sets until it passes local checks. Locally valid candidates are admitted only if they improve validation performance while preserving program semantics. All outcomes are stored in evolution memory to guide later revisions.

We evaluate \textsc{VeriSkill} across different verification tools, including Dafny, Frama-C, and VeriFast. \textsc{VeriSkill} consistently outperforms non-evolution and existing baselines of skill self-evolution, while ablation studies confirm the contribution of its core components. Moreover, the evolved skills retain their effectiveness across different agent frameworks and LLM backends. A case study further shows how \textsc{VeriSkill} turns a specific verifier failure into a reusable skill update, explaining its advantage over generic methods of skill self-evolution. 
To conclude, our primary contributions are as follows:

\begin{itemize}
    \item 
    We identify validated verification-responsibility patterns as the proper learning unit for reliable skill evolution in program verification.
    \item We propose the first self-evolution framework for program verification skills, which attributes reusable deficiencies, abstracts transferable knowledge, and iteratively validates revisions before admission.
    \item 
    We implement the framework as a prototype tool. Experiments show that \textsc{VeriSkill} consistently outperforms all baselines across multiple verification tools, agent frameworks, and LLM backends.
\end{itemize}

\section{Related Work}

\subsection{Skill Self-evolution}

Recent work learns reusable skills from agent experience. Trace2Skill consolidates trajectory-local lessons into transferable instructions~\cite{ni2026trace2skill}. SkillOpt and SkillOpt-Lite refine skill text from rollout feedback and retain revisions through validation~\cite{yang2026skillopt,shen2026skillopt}. EvoSkill analyzes execution failures and preserves effective skill folders through validation~\cite{alzubi2026evoskill}, while SkillCAT contrasts successful and failed trajectories before replaying candidate revisions~\cite{chen2026skillcat}. CoEvoSkills co-evolves a surrogate verifier to provide revision signals~\cite{zhang2026coevoskills}, whereas AutoSkill derives and reuses skills from dialogue and interaction traces for lifelong learning~\cite{yang2026autoskill}. These methods mainly rely on task outcomes or learned verification; \textsc{VeriSkill} instead uses formal-verifier evidence to attribute skill-responsible failures and admits revisions only under semantic-preservation constraints.

\subsection{LLM-Assisted Program Verification}

Recent studies investigate how LLMs can assist program-verification tasks. DafnyBench establishes a benchmark protocol for proof-hint generation with iterative verifier feedback~\cite{loughridge2024dafnybench}, while FM-Bench decomposes program verification into multiple subtasks across several specification languages to evaluate distinct verification capabilities~\cite{cao2025informal}. AxDafny further demonstrates that iterative generation and verification can be orchestrated through an agentic workflow~\cite{breen2026axdafny}.

A second line of work synthesizes specifications and annotations through verifier-guided refinement. SpecGen generates pre- and postconditions~\cite{ma2024specgen}, AutoSpec iteratively refines specifications according to verifier feedback~\cite{wen2024enchanting}, and Laurel first localizes a missing Dafny assertion from verifier diagnostics before generating it~\cite{mugnier2025laurel}. Preguss extends this paradigm to large C programs by combining static-analysis-guided decomposition with verifier-driven refinement of interprocedural specifications~\cite{wang2026tale}.

Other systems focus on invariant synthesis and structured proof repair. Lemur integrates LLM-generated lemmas and invariants into deductive verification through backtracking~\cite{wu2024lemur}, LaM4Inv couples LLM generation with bounded model checking~\cite{wu2024llm}, and Loopy applies Houdini-style filtering to candidate invariants~\cite{kamath2023finding}. AutoVerus further orchestrates specialized agents for different verifier error types and performs staged proof refinement~\cite{yang2025autoverus}, resembling \textsc{VeriSkill}'s use of diagnostic signatures to organize verification failures. However, across these systems, verifier feedback is primarily consumed within individual attempts to repair a specific artifact. The resulting procedural knowledge is not externalized as a persistent skill reusable across tasks, verifiers, and model backbones. \textsc{VeriSkill} fills this gap by converting validated verifier feedback into a portable skill artifact.

\begin{figure*}[t]
\centering
\includegraphics[width=\textwidth]{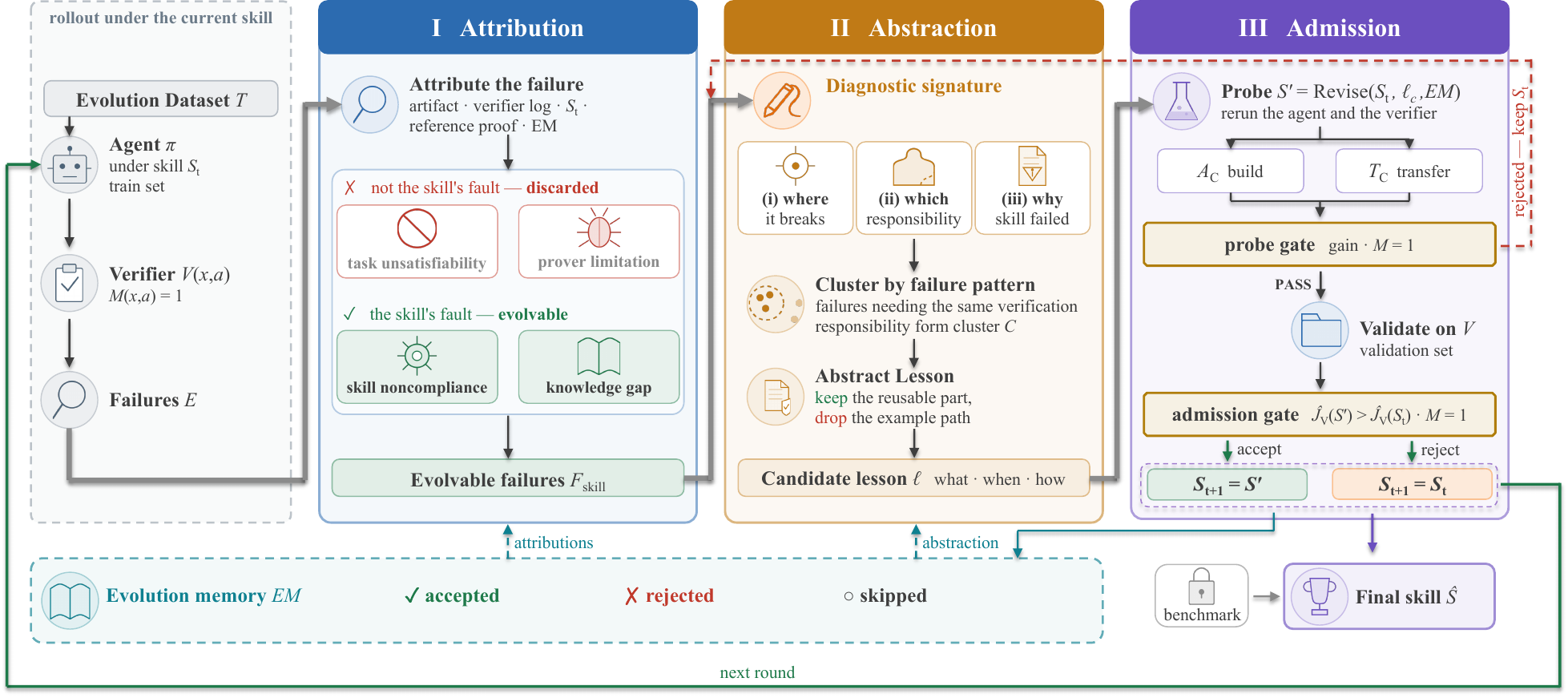}
\caption{Overview of \textsc{VeriSkill}. The framework (1) attributes failures and retains only skill-responsible cases, (2) clusters compatible diagnostic signatures to abstract reusable lessons, and (3) iteratively refines candidate skills through executable validation.}
\label{fig:overview}
\end{figure*}

\section{Problem Formulation}

Our goal is to evolve a verification skill into a more effective one that improves the agent's performance on program-verification tasks. 
Specifically, let $\pi$ denote the agent's solution-generation policy. Then $\pi(\cdot\mid x,S)$ is the conditional distribution over candidate solutions given task $x$ and skill $S$, where $\cdot$ denotes the candidate-solution argument of the distribution. Given a skill $S$, a distribution $\mathcal{D}$ over verification tasks, a program-verification task $x\sim\mathcal{D}$, and a candidate solution $a\sim\pi(\cdot\mid x,S)$, the expected verification success rate of $S$ is defined as
\begin{equation}
J(S)=
\mathbb{E}_{x\sim\mathcal{D}}
\mathbb{E}_{a\sim\pi(\cdot\mid x,S)}
[V(x,a)],
\end{equation}
where $V(x,a)\in\{0,1\}$ indicates whether the candidate solution passes the verifier. Therefore, the objective of skill self-evolution is to find a better skill:
\begin{equation}
S^\star \in \arg\max_S J(S).
\end{equation}
At the same time, any candidate solution must satisfy the semantic-preservation constraint:
\begin{equation}
M(x,a)=1,
\end{equation}
where $M(x,a)$ indicates that the candidate solution should not modify the original executable program. 

During this process, a verification failure cannot be directly equated with a defect in the skill. A failure may arise from the program or annotation itself, from the limits of the verification tool's solving capabilities, or from the agent failing to follow the existing skill guidance. If a skill is modified directly based on failure evidence, the system might encode problems that cannot be solved by skill updates, or accidental execution errors, into the skill, potentially pushing the skill self-evolution in the wrong direction.

Therefore, reliable skill self-evolution for program verification requires three stages.   
First, it performs responsibility attribution to identify failures that can serve as valid signals for skill updates. Second, it conducts fine-grained diagnosis of skill defects and abstracts reusable lessons from failure clusters. Finally, it uses executable validation and an admission mechanism to determine whether a candidate revision truly improves the complete skill. Together, these requirements transform skill optimization from ``modify whenever a failure is observed'' into ``admit only updates that have been attributed, abstracted, and validated''.

\section{VeriSkill}



This section presents the proposed skill-evolution framework, with three stages detailed in the following subsections. 
We also illustrate the overall workflow in Figure~\ref{fig:overview} and provide the corresponding pseudocode in Algorithm~\ref{alg:veriskill}.



\subsection{Stage 1: Failure Attribution and Filtering}

The first stage determines whether a verification failure should trigger skill revision. Its purpose is to distinguish failures attributable to the skill from those caused by the task or verification environment. Given a failed task, \textsc{VeriSkill} performs responsibility attribution by jointly considering the generated result, verifier output, skill, human-verified reference solution, and evolution memory. It classifies the failure into the following four categories.


\begin{itemize}
\item \emph{Task unsatisfiability}: The source program and annotations do not support the verification objective. 
\item \emph{Prover limitation}: The proof obligation is reasonable, but the current prover lacks the capability to discharge it. 
\item \emph{Skill noncompliance}: The skill provides relevant guidance, but the agent fails to follow it. 
\item \emph{Skill knowledge gap}: 
The skill lacks the reusable procedural knowledge required for the verification task.
\end{itemize}
The first two categories are excluded from skill evolution, whereas the latter two are retained as skill-related failures.

\textsc{VeriSkill} also consults historical attribution evidence in evolution memory. Records identifying similar failures as task unsatisfiability, prover limitation, or outcomes of invalid revisions help prevent repeated misattribution. Based on the current evidence and this history, \textsc{VeriSkill} retains only skill-related failures that are likely to be improved through textual updates. These evolvable failures form the basis for lesson abstraction and revision generation in the subsequent stages.

\subsection{Stage 2: Lesson Abstraction via Pattern-Level Defect Clustering}

After Stage 1 identifies skill-related failures, Stage 2 determines how the skill should be revised without generating a patch directly from an individual example. \textsc{VeriSkill} extracts diagnostic signatures, clusters failures with shared patterns, abstracts their common verification responsibility, and generates candidate lessons using evolution memory.


\textsc{VeriSkill} first extracts a diagnostic signature for each failure. The diagnostic signature contains the following three components. 
\begin{itemize}
\item \emph{Proof-obligation localization}: Identify where the proof fails, such as invariant initialization, invariant preservation, memory safety, or termination. 
\item \emph{Verification-responsibility abstraction}: Identify the reusable proof responsibility missing from the candidate solution, such as establishing entry facts, preserving an invariant, or supplying an intermediate lemma.
\item \emph{A skill failure mechanism}: 
Identifies how the skill text causes the failure, such as missing, unclear, overly abstract, or difficult-to-locate guidance.
\end{itemize}

Using these signatures, \textsc{VeriSkill} groups failures with a shared structure into failure-pattern clusters. It then abstracts the common verification responsibility within each cluster as its aggregate diagnosis. Reasoning over a cluster allows \textsc{VeriSkill} to capture semantically equivalent proof responsibilities across syntactically different solutions and avoids overfitting a revision to a single example.


When abstracting verification responsibility, \textsc{VeriSkill} refers to human-verified reference artifacts, but does not directly turn them into lessons. A reference solution typically contains two kinds of information: reusable verification responsibilities for the same type of task, and the specific path used for the current example, such as a particular invariant form, assertion location, lemma expression, or constant choice. \textsc{VeriSkill} retains only the former and removes the latter, thereby avoiding the hard-coding of example-level outputs into the skill.

To better leverage historical experience when reasoning about strategies of skill self-evolution, \textsc{VeriSkill} also retrieves records from evolution memory for the current failure pattern, including prior attributions, abstracted responsibilities, lessons, validation outcomes, and applicability conditions. Accepted revisions provide reusable procedural knowledge as guidance, whereas rejected or skipped revisions prevent repeated ineffective or unsafe updates. With memory, lesson generation becomes a pattern-level update constrained by historical validation evidence rather than a local induction from the current failed examples.

Finally, \textsc{VeriSkill} generates candidate lessons based on the failure pattern, shares verification responsibility, and evolution memory. A candidate lesson must specify its applicability conditions, concrete proof steps, and inapplicable cases, while avoiding task IDs, file names, concrete variable names, or constants. The resulting lesson is thus not an example-level patch, but reusable procedural knowledge for a class of similar failure patterns.

\begin{algorithm}[t]
\caption{VeriSkill Evolution Loop}
\label{alg:veriskill}
\begin{algorithmic}[1]
\REQUIRE Current skill $S$, evolution tasks $T$, validation tasks $\mathcal V$, evolution memory $EM$
\LOOP
    \STATE \textit{// Stage 1: Failure attribution and filtering}
    \STATE $E \leftarrow \mathrm{ExecuteAndCollect}(S,T)$
    \STATE $F \leftarrow \mathrm{AttributeFailures}(E,S,EM)$
    \STATE $F_{\mathrm{skill}} \leftarrow \mathrm{FilterEvolvable}(F,EM)$
    \IF{$F_{\mathrm{skill}}=\emptyset$}
        \STATE \textbf{break}
    \ENDIF
    \STATE \textit{// Stage 2: Lesson abstraction via pattern-level defect clustering}
    \STATE $\mathcal{C} \leftarrow \mathrm{DiagnoseAndCluster}(F_{\mathrm{skill}},S,EM)$
    \STATE $C \leftarrow \mathrm{SelectCluster}(\mathcal{C},EM)$
    \STATE $(A_C,T_C) \leftarrow \mathrm{SplitAttributionTransfer}(C)$
    \STATE $K \leftarrow \mathrm{AbstractResponsibility}(A_C,S,EM)$
    \IF{$K=\emptyset$}
        \STATE $EM \leftarrow \mathrm{RecordSkipped}(EM,C)$
        \STATE \textbf{continue}
    \ENDIF
    \STATE \textit{// Stage 3: Executable validation and admission}
    \REPEAT
        \STATE $\ell \leftarrow \mathrm{BuildLesson}(K,S,EM)$
        \STATE $S' \leftarrow \mathrm{ConstructCompleteCandidate}(S,\ell,EM)$
        \STATE $Z \leftarrow \mathrm{EvaluateLocalCandidate}(S,S',A_C,T_C)$
        \STATE $EM \leftarrow \mathrm{RecordLocalOutcome}(EM,C,\ell,S',Z)$
    \UNTIL{$\mathrm{LocalPasses}(Z) \lor \mathrm{BudgetExhausted}(C)$}
    \IF{$\neg\mathrm{LocalPasses}(Z)$}
        \STATE \textbf{continue}
    \ENDIF
    \IF{$\widehat{J}_{\mathcal V}(S')>\widehat{J}_{\mathcal V}(S) \land M(x_i,a_i^{S'})=1,\;\forall x_i\in\mathcal V$}
        \STATE $S \leftarrow S'$
        \STATE $EM \leftarrow \mathrm{RecordAccepted}(EM,C,\ell,S',Z,\mathcal V)$
    \ELSE
        \STATE $EM \leftarrow \mathrm{RecordRejected}(EM,C,\ell,Z,\mathcal V)$
    \ENDIF
\ENDLOOP
\RETURN $S$
\end{algorithmic}
\end{algorithm}

\subsection{Stage 3: Executable Validation and Admission}

The goal of the third stage is to iteratively refine the candidate skill so that it transfers to unseen cases with the same failure pattern before admission.
Given the current skill $S_t$, a candidate lesson $\ell_c$, and evolution memory $EM$, \textsc{VeriSkill} first constructs a candidate skill:
\begin{equation}
  S'
  =
  \mathrm{Revise}(S_t,\ell_c,EM).
\end{equation}
This is a controlled revision rather than a textual append. The lesson is integrated into the appropriate part of the skill according to its applicability scope and revision type. For example, the candidate may rewrite an unclear process, add conditional guidance for an uncovered case, or turn an abstract principle into explicit execution steps, while preserving unrelated sections and the original task contract.

\textsc{VeriSkill} then reruns the agent and verifier with the candidate skill $S'$ on attribution cases and held-out transfer cases from the same failure pattern. Let $\mathcal{A}_c$ denote the attribution cases used to analyze the failure pattern and construct the lesson, and let $\mathcal{T}_c$ denote same-pattern disjoint transfer cases that were not involved in lesson construction. Their local effects are defined as
\begin{equation}
  \begin{aligned}
  \widehat{\tau}_{\mathrm{attr}}(S')
  &=
  \frac{1}{|\mathcal A_c|}
  \sum_{x_i\in\mathcal A_c}
  \left[
  V(x_i,a_i^{S'})
  -
  V(x_i,a_i^{S_t})
  \right],\\
  \widehat{\tau}_{\mathrm{trans}}(S')
  &=
  \frac{1}{|\mathcal T_c|}
  \sum_{x_i\in\mathcal T_c}
  \left[
  V(x_i,a_i^{S'})
  -
  V(x_i,a_i^{S_t})
  \right].
  \end{aligned}
  \label{eq:local-effect}
\end{equation}
Here, $\widehat{\tau}_{\mathrm{attr}}$ checks whether the candidate skill repairs the failures that motivated the revision, while $\widehat{\tau}_{\mathrm{trans}}$ checks whether the same revision transfers to unseen cases with the same failure pattern. Local testing also rejects pass-to-fail regressions or modifications to the original executable code. If the candidate fails either local test, \textsc{VeriSkill} uses the resulting repair and regression evidence to refine the lesson and revise the skill again for the same failure pattern. This revise--test loop continues until a candidate passes both local tests or the failure-pattern budget is exhausted.

Only a locally valid candidate proceeds to the disjoint validation set $\mathcal{V}$. The success rate on the validation set is defined as
\begin{equation}
  \widehat{J}_{\mathcal{V}}(S)
  =
  \frac{1}{|\mathcal V|}
  \sum_{x_i\in\mathcal V}
  V(x_i,a_i^S),
\end{equation}
where $a_i^S$ denotes the candidate artifact generated by the agent for task $x_i$ under the guidance of skill $S$. A candidate that passes frozen validation is admitted directly. Specifically, \textsc{VeriSkill} requires the candidate skill to strictly improve the validation-set success rate while preserving the integrity of all validation tasks:

\begin{equation}
    S_{t+1}
    =
    \begin{cases}
    S',
    &
    \begin{aligned}[t]
    &\widehat{J}_{\mathcal V}(S')>\widehat{J}_{\mathcal V}(S_t)
      \land M(x_i,a_i^{S'})=1,\\
    &\forall x_i\in\mathcal V,
    \end{aligned}\\
    S_t,
    &
    \text{otherwise}.
    \end{cases}
    \label{eq:admission}
\end{equation}


An admitted skill becomes the evolution target for the next round. Meanwhile, every candidate revision, whether accepted, rejected, or skipped, is recorded in evolution memory with its failure pattern, attribution, lesson, validation outcome, and applicability conditions. Later rounds reuse these records to guide verification-responsibility attribution and lesson generation while avoiding revisions already shown to be ineffective or risky.

Thus, through local probing, iterative validation, and memory updates, \textsc{VeriSkill} turns candidate lessons into skill self-evolution constrained by executable verification evidence.

\begin{table*}[t]
\centering
\small
\renewcommand{\arraystretch}{1.15}
\setlength{\tabcolsep}{5pt}
\begin{tabular}{lccccccccc}
\toprule
Verifier & No Skill & LLM Skill & Human Skill & Skill-Creator & EvoSkill & AutoSkill & SkillOpt-Lite & \textbf{\textsc{VeriSkill}} & \textbf{$\Delta$}\\
\midrule
\multicolumn{10}{l}{\emph{Agent framework: Claude Code~~$\bullet$~~LLM backbone: Opus~4.8}} \\
\midrule
Dafny             & 14.0 & 21.3 &  7.7 & 39.3 & 32.3 & 41.0 & \underline{42.3} & \textbf{57.3} & \textbf{+43.3} \\
Frama-C           & 56.9 & 47.1 & 58.2 & 58.2 & 54.2 & 60.8 & \underline{65.4} & \textbf{74.5} & \textbf{+17.6} \\
VeriFast          & 38.0 & 32.0 & 24.3 & 58.3 & 53.0 & 62.7 & \underline{78.3} & \textbf{84.0} & \textbf{+46.0} \\
\midrule
\multicolumn{10}{l}{\emph{Agent framework: Codex~~$\bullet$~~LLM backbone: GPT~5.6~Sol}} \\
\midrule
Dafny             & 17.0 & 27.3 & 12.3 & 21.7 & 30.3 & 35.7 & \underline{62.7} & \textbf{66.0} & \textbf{+49.0} \\
Frama-C           & 57.5 & 38.6 & 59.5 & 56.2 & 55.6 & 60.1 & \underline{66.7} & \textbf{83.0} & \textbf{+25.5} \\
VeriFast          & 41.7 & 44.0 & 35.0 & 63.0 & 58.0 & 67.7 & \underline{76.0} & \textbf{93.0} & \textbf{+51.3} \\
\bottomrule
\end{tabular}
\caption{Verifier PASS rate (\%) on three verification tracks under two agent configurations. \textbf{Bold} marks the best score per row; \underline{underline} marks the second-best. The \textbf{$\Delta$} column reports the absolute improvement of \textsc{VeriSkill} over the \emph{No Skill} baseline.}
\label{tab:main}
\end{table*}

\section{Experimental Setup}

\paragraph{Datasets and Benchmarks.}
We evaluate on three verification tools with disjoint datasets and benchmarks.

\noindent \textit{Dafny}: Datasets and benchmarks are constructed from DafnyBench~\cite{loughridge2024dafnybench}.
Because DafnyBench comprises a large set of Dafny programs and is not inherently intended as a benchmark for C-to-Dafny translation, we sample 200 tasks, manually construct plain-C inputs paired with hidden Dafny references, and split them 4:1:5 into training, validation, and benchmark.

\noindent \textit{Frama-C}: Datasets are built by searching GitHub with Frama-C/WP and ACSL keywords, inspecting candidate repositories in descending relevance order, and selecting four relevant repositories as data sources~\cite{framacSnapshotRepo,blanchardTutorielWpRepo,framacOpenSourceCaseStudiesRepo,acslByExampleRepo}.
We parse function-level examples, strip ACSL annotations to form plain-C inputs, filter and deduplicate verifier-passing references, and manually check the resulting pairs. Two experts with formal-verification experience cross-validate the correctness of the annotations. This yields 85 C-to-ACSL evolution tasks split 68/17 for training and validation. The benchmark is the Frama-C-Problems benchmark~\cite{framaCProblemsRepo}.

\noindent \textit{VeriFast}: Since little existing C-to-VeriFast benchmark is available, we construct both the datasets and benchmarks from GitHub sources~\cite{verifastRepo,verifastExamplesIndex}, following the same procedures as Frama-C. After function parsing, VeriFast-annotation stripping, verifier filtering, deduplication, and manual checking, we obtain 200 C-to-VeriFast tasks, split them 4:1:5 into training, validation, and benchmark.

\paragraph{Agents and baselines.}
The main comparison uses two agent/backbone pairs: \textbf{Claude Code}~\cite{claudeCodeDocs} with \textbf{Opus~4.8}, and \textbf{Codex}~\cite{openaiCodexDocs} with \textbf{GPT~5.6~Sol}.
For cross-model transfer, we run GLM-5.2, Kimi-K2.7, Qwen3.7-Max, DeepSeek-V4-Pro, and Qwen3-Coder through the same Terminus terminal-agent wrapper~\cite{terminusAgent}, keeping prompts, verifier commands, and task environments fixed while varying only the backend model.

We compare \textsc{VeriSkill} against $7$  baselines.
\textbf{No Skill} gives the agent only the task prompt, while \textbf{LLM Skill} adds a one-pass generated verifier skill following the practice~\cite{yan2026openskill}.
\textbf{Human Skill} uses GitHub-sourced expert skills for Dafny and Frama-C, both validated by a formal-verification expert with seven years of experience; for VeriFast, where little suitable public skill was found, the same expert authored the skill.
\textbf{Skill-Creator} is Anthropic's official skill-authoring framework, which drafts reusable skill folders and iteratively revises them from self-graded feedback~\cite{anthropic2026claudeskills}.
\textbf{EvoSkill} maintains a Pareto frontier of agent programs and mutates selected frontier members by proposing new skill folders from sampled failure batches~\cite{alzubi2026evoskill}.
\textbf{AutoSkill} derives reusable skills from dialogue and interaction traces~\cite{yang2026autoskill}.
\textbf{SkillOpt-Lite} evolves skill text through trajectory exploration and validation gating~\cite{shen2026skillopt}.
All methods receive the same inputs, verifier settings, and target-agent configuration.

\paragraph{Metrics.}
Following patch-correctness assessment in automated program repair and execution-based evaluation in program translation~\cite{le2018overfitting,ye2021automated,roziere2020unsupervised}, our primary metric is \textbf{PASS}, defined as the percentage of benchmark tasks for which the generated artifact is accepted by the target verifier and also passes semantic-preservation checks.
The semantic-preservation check rejects artifacts that edit executable behavior or introduce verification-bypass constructs.
For annotation-only tracks, this requires the executable C projection to match the input after erasing annotations, comments, and whitespace.
For C-to-Dafny, the check compares aligned callables, control flow, state updates, returns, and assertions across the source and generated artifacts.
For every evaluation setting, we run each method three times and report the average as the final result.

\section{Experimental Results}

To provide a comprehensive evaluation, we conduct extensive experiments covering baseline comparisons, ablation studies, and cross-model transferability. Because each experiment requires repeated LLM-agent execution and formal verification over multiple evolution iterations, the complete evaluation incurred approximately US$\$15K$ in API costs.

\subsection{Main Results}

To evaluate whether \textsc{VeriSkill} produces a stronger skill than other baselines, we evaluate all methods on three verification tracks under two agent/backbone configurations and report the three-trial mean PASS rate in Table~\ref{tab:main}.

\textsc{VeriSkill} achieves the best score in all six verifier--agent configurations. Against \emph{No Skill}, it improves PASS by \textbf{+43.3}, \textbf{+17.6}, and \textbf{+46.0}~pp on Dafny, Frama-C, and VeriFast with Claude Code/Opus~4.8, and by \textbf{+49.0}, \textbf{+25.5}, and \textbf{+51.3}~pp with Codex/GPT~5.6~Sol. Compared with the strongest competing baseline in each row, \textsc{VeriSkill} still adds \textbf{+15.0}, \textbf{+9.1}, and \textbf{+5.7}~pp under Claude, and \textbf{+3.3}, \textbf{+16.3}, and \textbf{+17.0}~pp under GPT. Thus, the gains exceed prompt strength alone and remain above the strongest baseline SkillOpt-Lite.

Overall, evolved skills are clearly stronger than non-evolved skills: one-pass LLM Skill and Human Skill are inconsistent, sometimes falling below No Skill, whereas methods of skill self-evolution usually provide larger gains. However, these generic methods of skill self-evolution still lag behind \textsc{VeriSkill}, suggesting that program verification benefits from a customized evolution loop built around responsibility attribution, lesson abstraction, and executable validation.

\subsection{Ablation Study}

To explore which parts of \textsc{VeriSkill} are responsible for the outstanding performance, we isolate the three core mechanisms by removing one stage at a time on the Dafny track with Claude Code/Opus~4.8, while keeping the remaining pipeline and evaluation setting unchanged:
\begin{itemize}
    \item \textbf{w/o Responsibility Attribution}: every verification failure is treated as a reusable skill deficiency, regardless of alternative failure sources.
    \item \textbf{w/o Lesson Abstraction}: candidate guidance is derived from direct generated--reference differences rather than their proof functions.
    \item \textbf{w/o Executable Validation}: candidate skills skip the revise--test loop on attribution and transfer cases, proceeding directly to validation.
\end{itemize}

\begin{table}[t]
\centering
\small
\renewcommand{\arraystretch}{1.15}
\setlength{\tabcolsep}{7pt}
\begin{tabular}{lcc}
\toprule
\textbf{Setting} & \textbf{PASS} (\%) & \textbf{$\Delta$ vs.\ Full} \\
\midrule
Full \textsc{VeriSkill}              & \textbf{57.3} & --- \\
\midrule
\quad w/o Responsibility Attribution & 44.3 & $-13.0$ \\
\quad w/o Lesson Abstraction           & 49.0 & $-8.3$ \\
\quad w/o Executable Validation      & 36.0 & $-21.3$ \\
\bottomrule
\end{tabular}
\caption{Component ablation on the Dafny track (Opus~4.8). Each row removes one stage of \textsc{VeriSkill} while leaving the rest of the pipeline unchanged. \textbf{$\Delta$ vs.\ Full} reports the absolute drop from full \textsc{VeriSkill}.}
\label{tab:ablation}
\end{table}

\begin{table*}[t]
\centering
\footnotesize
\renewcommand{\arraystretch}{1.12}
\setlength{\tabcolsep}{2pt}
\begin{tabular}{p{0.11\textwidth}>{\centering\arraybackslash}p{0.43\textwidth}>{\centering\arraybackslash}p{0.43\textwidth}}
\toprule
 & \textbf{EvoSkill} & \textbf{\textsc{VeriSkill}} \\
\midrule
Source code & \multicolumn{2}{>{\centering\arraybackslash}p{0.86\textwidth}}{Null-pointer Judgement: \texttt{if (ptr == NULL) \{ abort(); \}}} \\
Skill update & \emph{Preserve failure-guard branches} & \emph{Do not fabricate branches for untranslatable failure guards} \\
Evaluation & Dafny PASS: 32.3\% & Dafny PASS: 57.3\% \\
\bottomrule
\end{tabular}
\caption{A Dafny case study on branch-count deficits.}
\label{tab:case-branch}
\end{table*}

Table~\ref{tab:ablation} shows that all three mechanisms are necessary, with executable validation contributing the largest margin. Removing the revise--test loop over attribution and transfer cases lowers PASS from 57.3 to 36.0, a \textbf{21.3}~pp drop. 
The loop uses cases from the same failure pattern to iteratively improve lesson accuracy before validation. Bypassing this process leaves lesson errors unresolved, limiting same-pattern generalization and propagating faulty guidance to later tasks.

Responsibility attribution is the second most important component, with a \textbf{13.0}~pp drop when removed. This indicates that treating every verifier failure as a skill defect injects noisy supervision into evolution, because some failures are caused by task unsatisfiability or prover limitations rather than missing reusable guidance. Removing lesson abstraction causes a smaller but still substantial \textbf{8.3}~pp drop, showing that abstract lessons act as reasoning guidance toward the correct revision direction rather than letting example-specific details dominate the update. Together, the ablations support the design of \textsc{VeriSkill}: attribution controls which failures become learning signals, abstraction derives reusable lessons, and executable validation iteratively improves lesson accuracy on same-pattern cases before global validation.

\subsection{Cross-Model Transferability}

\begin{figure}[t]
\centering
\includegraphics[width=\columnwidth]{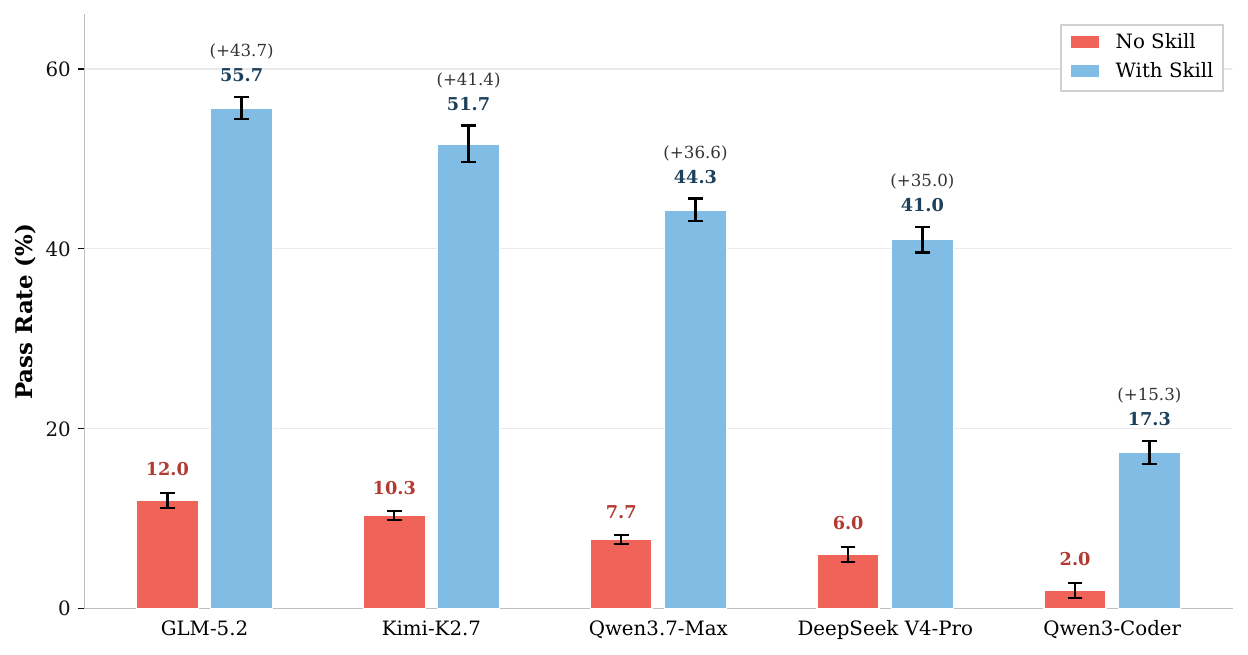}
\caption{Cross-model transferability under the Terminus agent framework.}
\label{fig:cross-model}
\end{figure}

We evaluate whether an evolved skill has cross-model transferability as reusable knowledge rather than remaining a model-specific prompt artifact. A skill is meant to externalize and stabilize procedural knowledge, so once useful verification guidance is extracted and written into the skill, it should transfer to other capable LLMs instead of remaining tied to the model used during evolution. To test this, we take the final skill evolved with Codex, deploy it unchanged in the Terminus agent framework, and compare each Terminus-backed LLM against its No Skill setting on the same Dafny benchmark.

The results in Figure~\ref{fig:cross-model} support this view. The Codex-evolved skill improves every Terminus-backed LLM, with gains ranging from +15.3 to +43.7~pp, even though no model-specific evolution is performed. This transferability indicates that \textsc{VeriSkill} does not merely tune agent- or LLM-specific behavior. It extracts verification procedures that can be reused as external knowledge by different models. Transferability is strongest for models that already have moderate no-skill performance, where the skill can redirect existing reasoning ability toward Dafny program construction. For weaker backends, the skill still helps, but the lower final PASS shows that reusable knowledge does not replace the model's ability to execute the verification strategy faithfully.

\section{Case Study}

We conduct a case study to explain why \textsc{VeriSkill} outperforms generic baselines of skill self-evolution. We compare with EvoSkill because it is conceptually closest to \textsc{VeriSkill}: both methods iteratively update skills from failures, but EvoSkill uses a generic failure-driven evolution loop rather than a program-verification-specific one.

Table~\ref{tab:case-branch} illustrates why \textsc{VeriSkill} improves over generic baselines of skill self-evolution. The same C fragment can lead to opposite skill updates: EvoSkill interpretation preserves the null-check branch to match source control flow, whereas \textsc{VeriSkill} recognizes that the branch is a host-level executability guard and should not be fabricated in Dafny when it carries no verified data-state behavior. This case explains why \textsc{VeriSkill} stands out: it does not merely react to failure signals, but interprets them through program-verification-specific attribution and lesson abstraction before updating the skill.

\section{Conclusion}


This paper presents \textsc{VeriSkill}, a self-evolution framework tailored to program verification that turns verification failures into targeted and trustworthy skill improvements. First, it performs responsibility attribution to retain only skill-responsible failures as evolution signals. Second, it clusters these failures by diagnostic signatures and abstracts reusable lessons that capture missing procedural guidance while removing instance-specific details. Finally, it iteratively refines candidate skills through executable validation and admits only revisions that improve verification performance while preserving program semantics. Evolution outcomes are stored in memory to guide subsequent revisions.

The experiments show that this program-verification-specific evolution framework \textsc{VeriSkill} consistently outperforms other methods across Dafny, Frama-C, and VeriFast. The ablation study confirms that responsibility attribution, lesson abstraction, and executable-validation checks each contribute to the final skill quality. The cross-model transferability experiment further shows that evolved skills can benefit other LLMs, supporting the view that \textsc{VeriSkill} extracts reusable verification knowledge rather than a model-specific prompt artifact. Finally, the case study explains this advantage concretely: \textsc{VeriSkill} does not merely react to failure signals, but interprets them through program-verification-specific attribution and lesson abstraction before updating the skill.


\bibliography{references}

\end{document}